\documentclass[conference]{IEEEtran}
\IEEEoverridecommandlockouts
\usepackage{cite}
\usepackage{amsmath,amssymb,amsfonts}
\usepackage{algorithmic}
\usepackage{graphicx}
\usepackage{textcomp}
\usepackage{bm}
\usepackage{url}
\usepackage{hyperref}

\usepackage{booktabs}

\begin{document}

\title{A surgical system for automatic registration, stiffness mapping and dynamic image overlay\thanks{This work has been funded through the National Robotics Initiative by NSF grant IIS-1426655.}
}

\author{\IEEEauthorblockN{ Nicolas Zevallos, Rangaprasad Arun Srivatsan, Hadi Salman, Lu Li,\\ Jianing Qian, Saumya Saxena, Mengyun Xu, Kartik Patath and Howie Choset}
\IEEEauthorblockA{Robotics Institute, Carnegie Mellon University,
5000 Forbes Avenue, Pittsburgh, PA 15213.\\
Email: (nzevallo, arangapr, hadis, lilu12, jianingq, saumyas, mengyunx, kpatath)@andrew.cmu.edu, choset@cs.cmu.edu}}

%
%
%

\maketitle

\begin{abstract}
In this paper we develop a surgical system using the da Vinci research kit (dVRK) that is capable of autonomously searching for tumors and dynamically displaying the tumor location using augmented reality. Such a system has the potential to quickly reveal the location and shape of tumors and visually overlay that information to reduce the cognitive overload of the surgeon. We believe that our approach is one of the first to incorporate state-of-the-art methods in registration, force sensing and tumor localization into a unified surgical system. First, the preoperative model is registered to the intra-operative scene using a Bingham distribution-based filtering approach. An active level set estimation is then used to find the location and the shape of the tumors.  We use a recently developed miniature force sensor to perform the palpation. The estimated stiffness map is then dynamically overlaid onto the registered preoperative model of the organ. We demonstrate the efficacy of our system by performing experiments on phantom prostate models with embedded stiff inclusions.  
\end{abstract}


\section*{Supplementary Material}
This paper is accompanied by a video: \href{https://drive.google.com/file/d/14wmEJ67NUNlQKs3VP1ZnodDg5a7YkwNW/view}{Click video Link}
\section{Introduction}
Robot-assisted minimally invasive surgeries (RMIS) are becoming increasingly popular as they provide increased dexterity and control to the surgeon while also reducing trauma, blood loss and hospital stays for the patient~\cite{jaydeeprmis09}. These devices are typically  teleoperated by the surgeons using visual feedback from stereo-cameras, but without any haptic feedback. This can result in the surgeon relying only on vision to identify tumors by mentally forming the correspondence between intra-operative view and pre-operative images such as CT scans/MRI, which can be cognitively demanding.

Automation of simple but laborious surgical sub-tasks and presenting critical information back to the surgeon in an intuitive manner has the potential to reduce the cognitive overloading and mental fatigue of surgeons~\cite{garg2016tumor}. This work leverages the recent advances in force sensing technologies~\cite{li2017development}, tumor localization strategies~\cite{ayvali2016using, ayvali2017utility, hadi2018trajectory}, online registration techniques~\cite{srivatsan2016estimating, srivatsanbingham2017} and augmented reality~\cite{kartikDynamic2017} to automate the task of tumor localization and dynamically overlay the information on  top of intraoperative view of the anatomy. 

\begin{figure}[ht!]
\includegraphics [width=0.5\textwidth]
{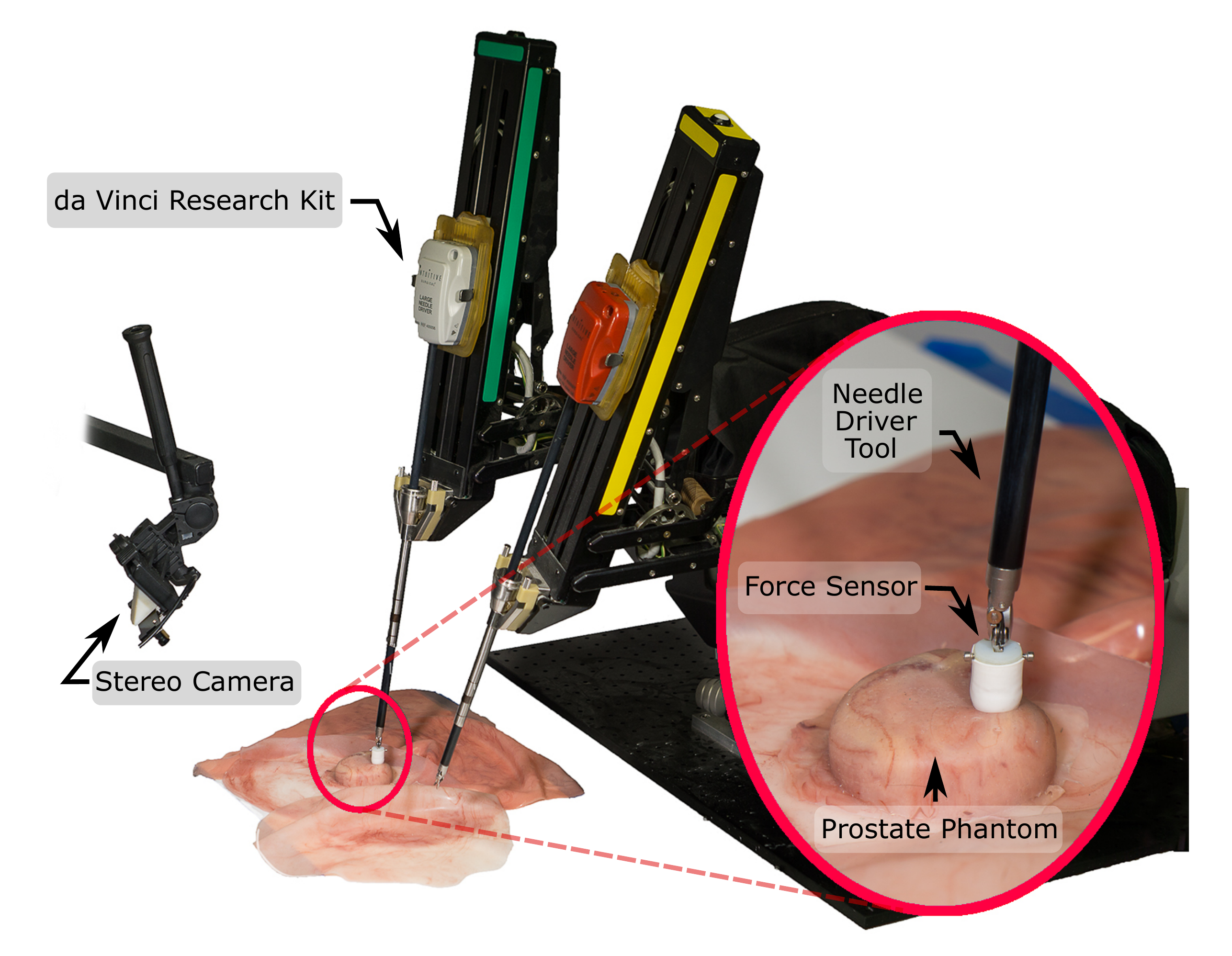}
\caption{Experimental setup showing the dVRK robot with a miniature force sensor attached to the end-effector. A stereo camera overlooks the workspace of the robot. A phantom prostate with embedded stiff inclusion is placed in the workspace of the robot. }
\label{fg:Setup}
\end{figure}

While the works in literature deal with force sensing~\cite{puangmali2008state, tiwana2012review}, tumor localization~\cite{ayvali2016using, garg2016tumor,ayvali2017utility,preetham2017online,  hadi2018trajectory} and graphical image overlays~\cite{shuhaiber2004augmented, teber2009augmented, su2009augmented, yamamoto2009tissue}, there is a gap in literature when it comes to systems that deal with all these issues at the same time. For example, Yamamoto~\emph{et al.}~\cite{yamamoto2009tissue} deal with tumor localization and visual overlay, but they assume the organ is flat and place the organ on a force sensing plate, which is not representative of a surgical scenario.  On the other hand, Garg~\emph{et al.}~\cite{garg2016tumor} use a palpation probe mounted on a da Vinci research kit (dVRK) tool~\cite{mckinley2015single}. However, they do not deal with registering the organ or visual overlay of the estimated stiffness map. This work, aims to bridge these shortcomings and present a unified system capable of addressing all the above mentioned issues at the same time. 

The system of Naidu~\emph{et al.}~\cite{naidu2017breakthrough} comes closest to our work. They  use a custom designed tactile probe to find tumors and visually overlay the tactile image along with the ultrasound images. The wide tactile array that they use, allows for imaging sections of the organ instead of obtaining discrete measurements, as in our case. This eliminates their need to develop sophisticated tumor search algorithms. However, as acknowledged by the authors~\cite{trejos2009robot}, it is not clear as to how their system would perform when using non-flat organs such as prostates and kidneys; since the tactile array cannot deform and confirm to the shape of the organ. Without performing registration, the image overlay would also be affected on non-flat organs.

The framework presented in this work is robot agnostic and modular in nature. We demonstrate the efficacy of the system by performing autonomous tumor localization on a phantom prostate model with embedded tumors using the dVRK (see Fig.~\ref{fg:Setup}. A miniature force sensor mounted at the tip of the dVRK needle driver tool~\cite{li2017development}  is used to  sense the contact forces. An active tumor search strategy~\cite{gotovos2013active, hadi2018trajectory} is used to localize the tumor. The estimated stiffness map is overlaid on a registered model of the anatomy and displayed in real-time on a stereo viewer.  

\section{Related Work}
\subsection{Force sensing for surgical applications}
The following survey papers report a number of devices that measure contact forces~\cite{puangmali2008state, tiwana2012review}. Some common drawbacks with many of the existing devices are: difficulty to sterilize, high cost , delicate components and lack of flexibility of form factor. Recently, our group has developed a miniature force sensor that uses an array of thin-film force-sensitive resistors (FSR)  with embedded signal processing circuits~\cite{li2017development}. The FSR sensor  is light weight, inexpensive, robust and has a flexible form factor. 
\subsection{Tumor search approaches}
The recent developments in force sensors have also resulted in a number of works that  automate mapping of the surface of the anatomy to reveal stiff inclusions. The different palpation strategies commonly used are: discrete probing motion~\cite{yamamoto2009tissue, nichols2015methods}, rolling motion~\cite{liu2010rolling} and cycloidal motion~\cite{goldman2013algorithms}. Some of these works direct the robot along a predefined path that scans the region of interest on the organ~\cite{yamamoto2009tissue, howe1995remote, srivatsan2016complementary }, while others adaptively change the grid resolution to increase palpation resolution around boundaries of regions with high stiffness gradients~\cite{goldman2013algorithms, nichols2015methods}.

Over the last two years, Bayesian optimization-based methods have gained popularity~\cite{ayvali2016using,  garg2016tumor,ayvali2017utility,preetham2017online}. These methods model the stiffness map using a Gaussian process regression (GPR) and reduce the exploration time by directing the robot to stiff regions. While the objective of most prior works is to find the high stiffness regions~\cite{ayvali2016using, garg2016tumor,ayvali2017utility}, our recent work on active search explicitly encodes finding the location and the shape of the tumor as its objective~\cite{hadi2018trajectory}. 
 
\subsection{Surgical registration and image overlay}
There is a rich literature of image overlay for minimally invasive surgeries~\cite{shuhaiber2004augmented}, including some works on usage of augmented reality in human-surgeries~\cite{marescaux2004augmented}. Often the image that is overlaid is a segmented preoperative model, and it manually placed in the intraoperative view~\cite{marescaux2004augmented, su2009augmented}. Very few works such as~\cite{teber2009augmented, haouchine2014towards}, deal with manual placement followed by automatic registration of the organ models. There are a number of registration techniques that have been developed for surgical applications; the most popular one being iterative closest point (ICP)~\cite{besl1992method} and its variants~\cite{rusinkiewicz2001}.

Probabilistic methods for registration have recently gained attention as they are better at handling noise in the measurements. Billings~\emph{et al.}~\cite{billings2015iterative} use a probabilistic matching criteria for registration, while methods such as~\cite{moghari2007point,srivatsan2016estimating} ( and the references therein) use Kalman filters to estimate the registration parameters. Our recent work reformulates registration as a linear problem in the space of dual quaternions and uses a Bingham filter and a Kalman filter to estimate the rotation and translation respectively~\cite{srivatsanbingham2017}. Such an approach has been shown to produce more accurate and fast online updates of the registration parameters.

While the above literature deals with registering preoperative models onto an intraoperative scene, there is very little literature that deals with overlaying stiffness maps on the preoperative models and updating the maps in real-time as new force sensing information is obtained. Real-time update is very important, because it gives the surgeon a better sense of what the robot has found and gives them insight into when to stop the search algorithm which is a subjective decision, as observed in~\cite{ayvali2017utility}. The works of Yamamoto~\emph{et al.}~\cite{yamamoto2009tissue} and Naidu~\emph{et al.}~\cite{naidu2017breakthrough} are exceptions and deal with dynamic overlaying of the stiffness image, but only onto flat organs. Their approaches do not generalize to the cases of non-flat organs such as kidneys or prostates that we consider in this work.

\section{Problem Setting and Assumptions}
We use an ELP stereo camera (model 1MP2CAM001) overlooking the workspace of a dVRK~\cite{kazanzides2014open}. A custom fabricated prostate phantom (made using \href{https://www.smooth-on.com/products/ecoflex-00-10/}{Ecoflex 00-10}) embedded with a plastic-disc to mimic a stiff tumor, is used for experimental validation.  

Given an \emph{a priori} geometric model of an organ, the measurements of the tool tip positions and associated contact forces, and stereo-camera images of the intraoperative scene, our goal is to (i) register the camera-frame, robot-frame and model-frame to each other, (ii) estimate the stiffness distribution over the organ's surface, and (iii) overlay the estimated stiffness distribution on the registered model of the organ and display it back to the user.

We make the following assumptions in this work:
\begin{itemize}
\item The shape of the organ never deforms globally but instead experiences local deformations only due to tool-interaction.
\item The tool-tip pose can be obtained accurately from the robot kinematics.
\item The forces applied by the tool are within the admissible range ( $<10$N) in which the organ only undergoes a small deformation ($<8$mm) that allows it to realize its undeformed state when the force is removed.
\item The stiff inclusion is located relatively close to the tissue surface, so that it can be detected by palpation.
\end{itemize}

\section{System Modeling and Experimental Validation}
\begin{figure}[ht!]
\includegraphics [width=0.5\textwidth]
{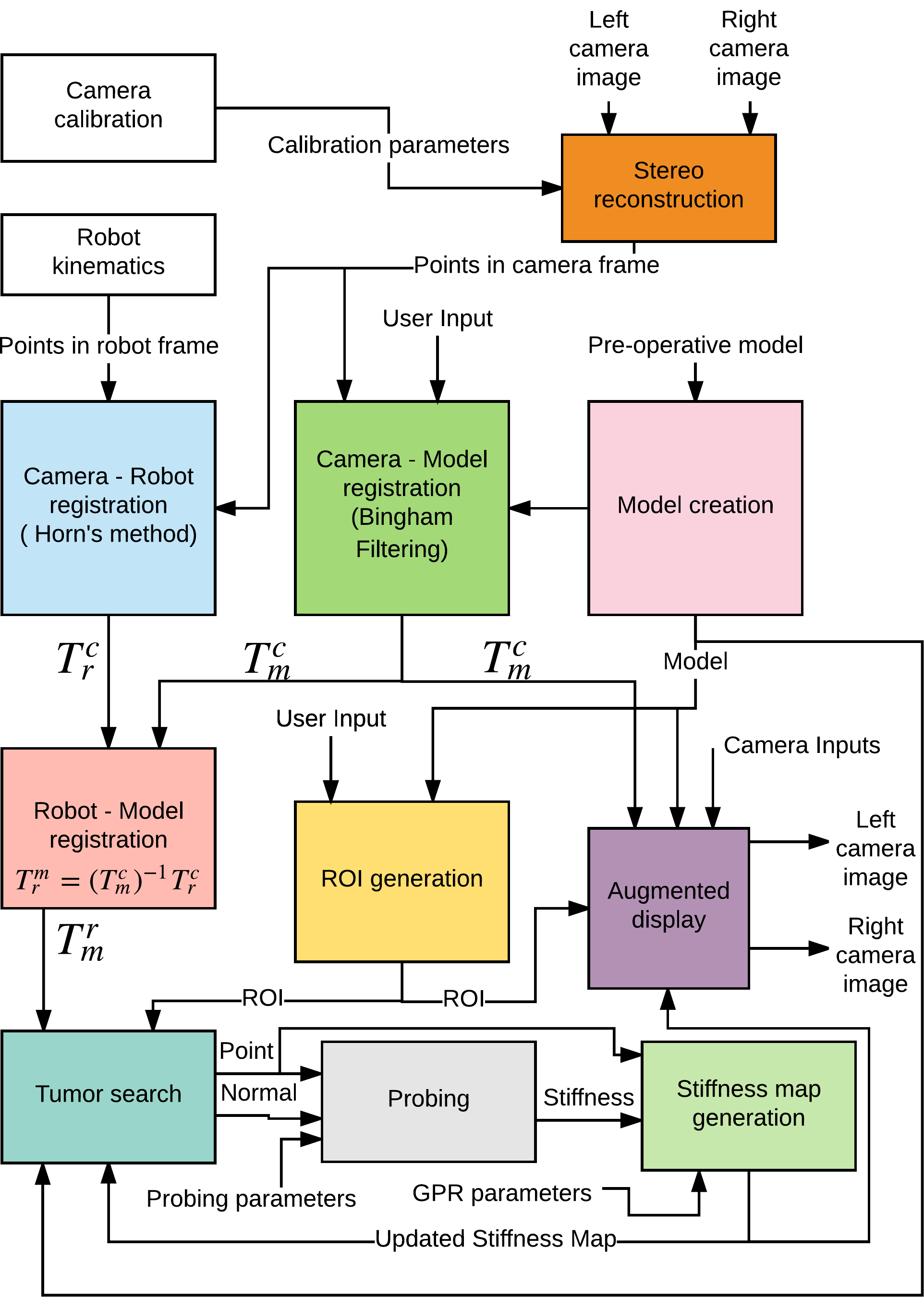}
\caption{Flowchart showing all the modular components of our system. Some of the modules such as camera calibration, stereo reconstruction, model creation, and camera-robot-model registrations are implemented once before the start of the experiment, while the other modules are constantly run for the duration of the experiment. }
\label{fg:mainFlowChart}
\end{figure}
Fig.~\ref{fg:mainFlowChart} shows the flowchart of the entire system. Modules such as camera calibration, model generation and registration need to be run only once at the beginning of the experiment. On the other hand, the tumor search, probing, and augmented display modules are run in a loop until the user is satisfied with the result and halts the process. While the system is largely autonomous, user input is required in two steps: (i) Camera-model registration, to select the organ of interest in the view of the camera, (ii) selecting region of interest for stiffness mapping. The modularity of the system allows the user to choose any implementation for registration, force-sensing and tumor localization. The important modules of our system are discussed in detail in the following sections.

\subsection{Registering Camera and Robot Frames}The cameras are calibrated using standard  \href{http://wiki.ros.org/camera_calibration/Tutorials/StereoCalibration}{ROS calibration}.
The robot is fitted with a colored bead on its end effector that can be easily segmented from the background by hue, saturation, and value. Registration between the camera-frame and the robot-frame is performed by the user through a graphical user interface (GUI) that shows the left and right camera images and has sliders representing color segmentation parameters.

The robot is moved to a fixed set of six points. These points are chosen to cover a substantial amount of the robot's workspace, stay within the field of view in the camera, and not contain symmetries that would make registration difficult. We chose to use only six points after experiments showed that additional points failed to significantly decrease the root mean squared error (RMSE), as shown in Table~\ref{tb:rmse}. For each of the points, we perform a series of actions.

First, we move the robot to the specified location, then we process both the left and right images to find the centroid of the colored bead fitted to the robot. The centroid of the ball in pixels is found as the center of the minimum enclosing circle of the contour with the largest area. We repeat this for as many frames as are received over ROS in one second (in our case 15), and the centroid is then averaged over all frames to reduce the effect of noise in the image. The centroid is drawn onto both images in the GUI, allowing the user to evaluate the accuracy of the centroid estimation. The pixel disparity is calculated as the difference between the $x$ coordinates of the centroid in the left and right images. This disparity is fed into a stereo-camera model that ROS provides, to calculate a 3D point in the camera-frame.
 \setlength{\tabcolsep}{1mm} 
\begin{table}[htbp]
\caption{}
\label{tb:rmse}
\centering
\begin{tabular}{l c c c c c c }
\toprule
 Number of points & 5 & 6& 7 & 8 & 11 & 51 \\  \midrule
RMSE (mm) & 2.71 & 2.37 & 2.84 & 3.01 & 2.82 & 2.85 \\
 \bottomrule
\end{tabular}
\end{table}

Following this, we obtain six points in both the camera-frame and the robot-frame (using the kinematics of the robot). We use Horn's method~\cite{horn1987} to calculate the transformation $\bm{T}^c_m$ between the camera and the robot frames. This transformation is saved to a file and the calculated RMSE is displayed to the user. In addition, the robot's current position is transformed by the inverse of the calculated transformation and projected back into the pixel space of both cameras. Circles are drawn at these pixel positions in the left and right images in the GUI so that the user can visually confirm that the registration is successful and accurate.

\subsection{Registering Camera and Preoperative Model Frames}
\label{sc:Tcm}
\begin{figure}[ht!]
\includegraphics [width=0.5\textwidth]
{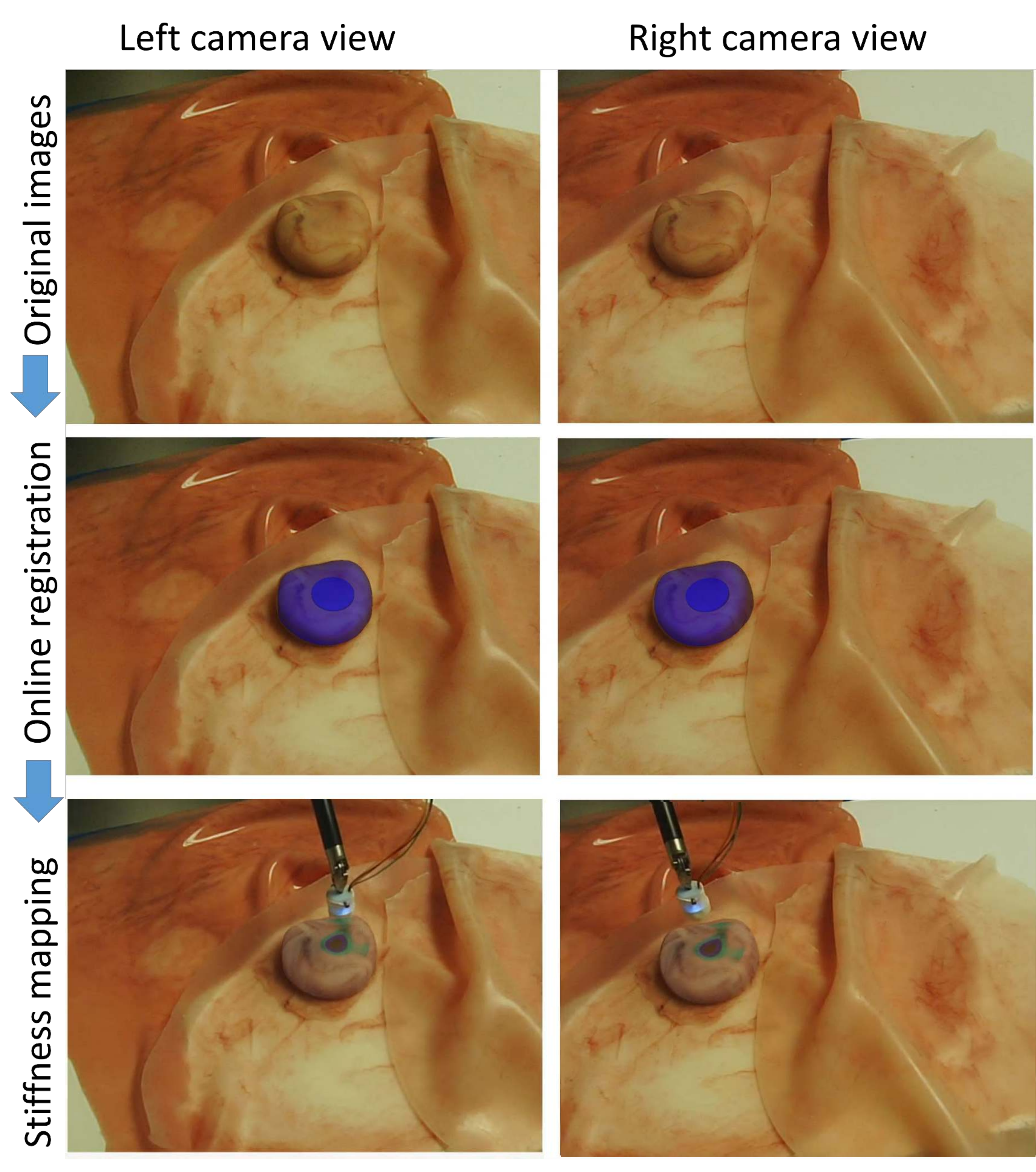}
\caption{Top row: Original left and right camera images. Middle row: Camera images with registered prostate model shown in semi-transparent blue. The tumor model is also shown to allow us to compare our stiffness mapping result. Bottom row: The robot probes the organ and records force-displacement measurements. The estimated stiffness map is then augmented on the registered model in this figure. Dark blue regions show high stiffness. Note that the stiffness map reveals the location and shape of the tumor.  }
\label{fg:registration}
\end{figure}

The transformation between camera-frame and model-frame, $\bm{T}^c_m$ is estimated by registering the reconstructed point cloud from stereo images with the preoperative model of the organ. The intraoperative scene as viewed by the stereo cameras is as shown in the top of Fig.~\ref{fg:registration}. A user manually selects the region containing the organ of interest. Following this the user can also further refine the selection using a graph cut-based image segmentation. 

A Bingham distribution-based filtering approach is used to automatically register the stereo point cloud to the preoperative model~\cite{srivatsanbingham2017}. The mean time taken to register is 2s and the RMS error is 1.5mm. The center row in Fig.~\ref{fg:registration} shows the registered model of the organ overlaid on the stereo views. Note how the pose of the registered model accurately matches the pose of the organ. In the same figure we also show the model of the tumor in the registered view to highlight how accurately the stiffness map estimates the location of the tumor (see bottom row of Fig.~\ref{fg:registration})

\subsection{Tumor Search and Stiffness Mapping}
The problem of tumor search is often posed as a problem of stiffness mapping, where the stiffness of each point on a certain organ is estimated, and regions with stiffness higher than a certain threshold are considered as regions of interest (tumors, arteries, etc.). The framework that we use for localizing tumors utilizes Gaussian processes (GP) to model the stiffness distribution combined with a GP-based acquisition function to direct where to sample next for efficient and fast tumor localization.

By using GP, we assume a smooth change in the stiffness distribution across the organ. Since every point on the organ's surface can be uniquely mapped to a 2D grid, the domain of search used is $X\subset {\rm I\!R^2}$. The measured force and position after probing the organ by the robot at $\bm{x}$ provides the stiffness estimation represented by $y$.


The problem of finding the location and shape of the stiff inclusions can be modeled as an optimization problem. However, an exact functional form for such an optimization is not available in reality. Hence, we maintain a probabilistic belief about the stiffness distribution and define a so called ``acquisition function", $\xi_{acq}$, to determine where to sample next. This acquisition function can be specified in various ways and thus our framework is flexible in terms of the choice of this acquisition function that is being optimized. Our prior works have considered various choices for the acquisition functions such as expectation improvement (EI), upper confidence bound (UCB), uncertainty sampling (UNC), active areas search (AAS) and active level sets estimation (LSE)~\cite{ayvali2016using,ayvali2017utility,hadi2018trajectory}. 

While our system is flexible to the choice of acquisition function, in this work we demonstrate tumor localization using LSE. LSE determines the set of points, for which an unknown function (stiffness map in our case) takes value above or below some given threshold level $h$. The mean and covariance of the GP can be used to define a confidence interval, 
\begin{equation}
Q_t(\bm x) = \left[\mu_t(\bm x) \pm \beta^{1/2} \sigma_t(\bm x)\right]
\end{equation} 
for each point $\bm x \in \bar X$. Furthermore, a confidence region $C_t$ which results from intersecting successive confidence intervals can be defined as,
\begin{equation}
C_t(\bm x) = \bigcap_{i=1}^t Q_i(\bm x).
\end{equation}
LSE then defines a measure of classification ambiguity $a_t(\bm x)$ defined as,
\begin{equation}
\label{eq:lseaqfxn}
a_t(\bm x) = \min\left\{max(C_t(\bm x))-h, h - \min(C_t(\bm x)) \right\}.
\end{equation}
LSE chooses sequentially queries (probes) at $\bm x_*$ such that,
\begin{equation}
\bm x_* = \operatorname*{arg\,max}_{\bm x\in X} {a_t(\bm x)}.
\end{equation}
For details on how to select the parameter $h$, we refer the reader to the work of Gotovos~\emph{et al.}~\cite{gotovos2013active}.

\subsection{Probing and Force Sensing}
We adopted a miniaturized Tri-axial sensor developed in~\cite{li2017development} onto the needle driver tool for the dVRK, to provide contact force measurements~(see Fig.~\ref{fg:Setup}). The force sensor is a Force-Sensitive-Resistor (FSR) based force-to-voltage transducer operating in thru-mode electrodes configuration. The design combines FSR array with a center mounted pre-load mechanical structure to provide a highly responsive measurement of contacting force and direction of the force vector. In this experiment, we electrically bridged the four sensing array elements on the force sensor, to provide improved sensitive force measurement along the normal direction of the sensor, since the dVRK can be  accurately oriented to probe along the local surface normal. In addition, we implemented online signal processing software in the sensor embedded controller, for analog signal amplification, filtering, automatic self-calibration, which is crucial step to improve sensor performance when using inexpensive force sensing materials such as 3M Velostat film from \href{https://www.adafruit.com/product/1361}{Adafruit}. 

\begin{figure}[ht!]
\includegraphics [width=0.5\textwidth]
{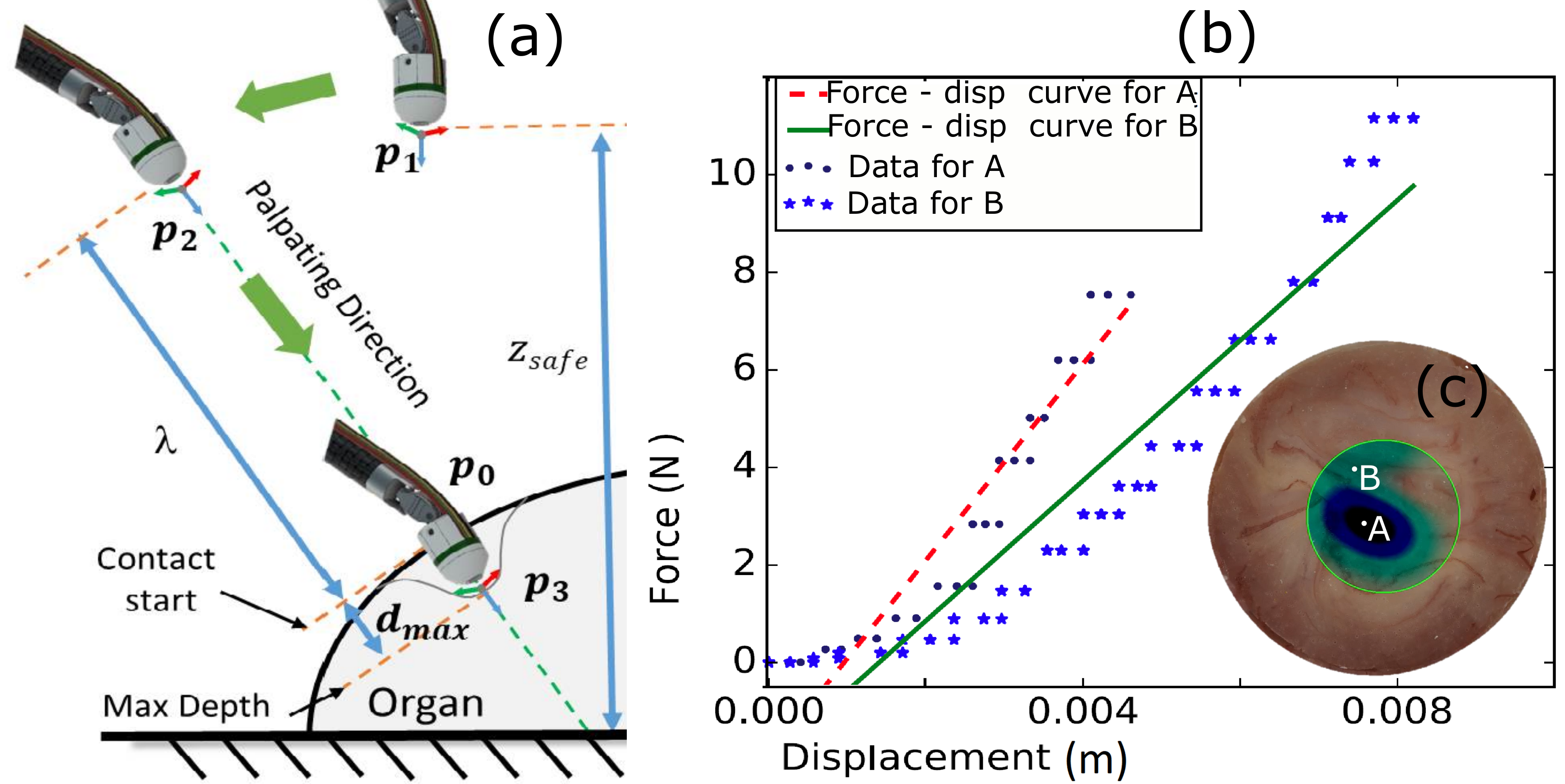}
\caption{(a) The various steps taken to probe a desired point along a desired normal direction as provided by the tumor search module. (b)The plot shows forces vs displacement for two sample points A and B on the surface of the organ. Note that the forces are limited to 10N and the displacement is also restricted to 8mm. RANSAC is used to find the best-fit line and the slope gives us an estimate of the stiffness at the probed location. (c) This 2D space forms a one-to-one mapping with the 3D surface of the organ. The green circle represents the user-defined ROI. The stiffness map is estimated in this ROI. Different shades of blue are used to represent the stiffness values. Point A is located in on a stiff region, while B is located on a soft region. The plot reveals the corresponding stiffness. }
\label{fg:StiffnessMapping}
\end{figure}

First, the robot is commanded to a safe position $\bm{p}_1$ which is at a known safe height $z_{safe}$ as shown in Fig.~\ref{fg:StiffnessMapping}(a). The robot is then commanded to move to position $\bm{p}_2$ which is at an estimated distance $\lambda$ from the desired probing point $\bm{p}_0$, along the normal to the surface at $\bm{p}_0$, $\bm{n}$ (see Fig.~\ref{fg:StiffnessMapping}(a)). While maintaining its orientation, the tool is commanded to move to position $\bm{p}_3=\bm{p}_2-(\lambda+d_{max})\bm{n}$. The force and position data are constantly recorded as the robot moves from $\bm{p}_2$ to $\bm{p}_3$. When the force sensor contacts the tissue surface, if the contact force exceeds a set threshold $F_{max}$ or if the probe penetrates more than a set depth $d_{max}$, the robot is no longer moved. This ensures that the probing does not hurt the patient or cause any damage to the robot. Following this we retract the robot to position $\bm{p}_2$ and then $\bm{p}_1$. Note that we do not record force and displacement data during the retraction process.
 
Next the recorded data is treated as input to the stiffness mapping algorithm similar to~\cite{srivatsan2016complementary}. There are two important steps of this algorithm: (i) baseline removal, (ii) stiffness calculation. Ideally, the force sensor reading should be zero when there is no contact between force sensor and the interest area. However, in reality there is always a small residue in the sensor readings even when there is no contact. Thus we find the mean sensor output value when the probe is at $\bm{p}_2$ and then subtract all the subsequent measurements from this baseline force. For stiffness calculation, we use a standard RANSAC algorithm to find the best fit line between the y-axis (force sensor data) and x-axis (displacement data).  As a result, the calculated regression coefficient indicates the changing rate of the contact force respect to a unit displacement, which can be used as the best approximation of stiffness value. Fig.~\ref{fg:StiffnessMapping}(b) shows the nearly linear variation of force with displacement, justifying the use of slope of the best fit line as an approximation for the stiffness.

\subsection{Dynamic Image Overlay}
The rendering of the overlays is done using the \href{https://www.vtk.org/}{Visualization Toolkit} (VTK).  Two virtual 3D cameras are created to match the real cameras using the results of camera calibration. The pre-operative model is placed in virtual 3D space according to the camera-to-organ registration, $\bm{T}^c_m$, and rendered as a polygonal mesh from the perspective of each camera. These two renders are overlaid onto live video from the left and right camera feeds as their backgrounds.
\begin{figure}[ht!]
\includegraphics [width=0.5\textwidth]
{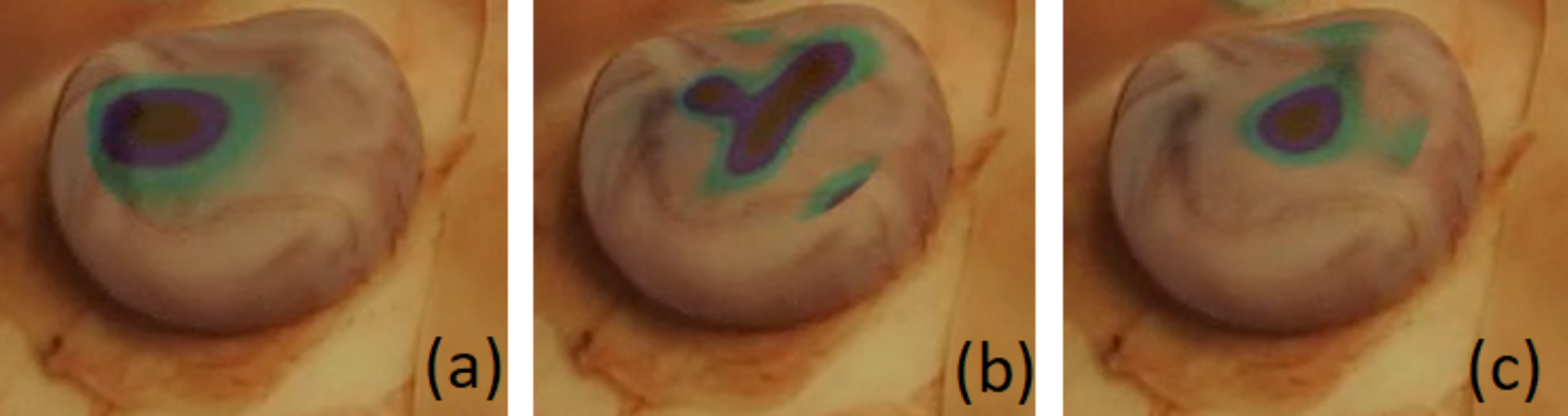}
\caption{The figures show the augmented stiffness map at various stages of probing.  The high stiffness regions are shown in darker shades of blue, while the low stiffness regions are in lighter shades of blue.(a) Result after a single probe, (b) result after 4 probings, (c) result after 10 probings.}
\label{fg:DynamicStiffnessMapping}
\end{figure}
These renderings are displayed in a GUI divided into three tabs. The first tab is for registration, which overlays the pre-operative model as described above and additionally allows the user to mask and segment the point cloud as described in Sec.~\ref{sc:Tcm}. It also provides buttons to start and stop model registration. The second tab allows the user to select a region of interest (ROI) defined in a 2D UV texture map that represents a correspondence between pixels on a 2D image to 3D coordinates on the surface of the pre-operative model  (see Fig.~\ref{fg:StiffnessMapping}(c)). The third tab overlays the pre-operative model over the camera feeds and allows the user to set the opacity of the overlay using a slider at the bottom of the window.

In addition, the renderings in the third tab add a texture to the rendered model. For this texture, the results of the tumor search are turned into a heat-map image representing relative stiffness in a user-specified region of interest (ROI) (see Fig.~\ref{fg:StiffnessMapping}(c)). This ROI is defined in 2D UV texture coordinates that represent a correspondence between pixels on a 2D image to 3D coordinates on the surface of the polygonal mesh. The heat-map image is broadcast over ROS and overlaid onto the pre-operative model's 2D texture image resulting in dark marks in high-stiffness areas while preserving texture details found in the pre-operative model's original texture (see Fig.~\ref{fg:StiffnessMapping}(c)). This 2D texture is then applied to the polygonal mesh using the UV map, resulting in a 3D overlay of the stiffness map onto the video feed from each camera. Fig.~\ref{fg:DynamicStiffnessMapping} shows the stiffness maps at various stages of probing, dynamically overlaid on the registered model of the organ. Note that the stiffness map clearly reveals the location and shape of the tumor which is shown in the middle row of Fig.~\ref{fg:registration}.

\section{Discussions and Future Work}
In this paper, we presented a system that unifies autonomous tumor search with augmented reality to quickly reveal the shape and location of the tumors while visually overlaying that information on the real organ. This has the potential to reduce the cognitive overload of the surgeons and assist them during the surgery. Our system demonstrates promising results in experimentation on phantom  silicone organs. 

While we demonstrate the task of stiffness mapping in this work, our system can be used to visually overlay pre-surgical plans, ablation paths, annotate important landmarks, etc. to aid the surgeon during the procedure. In our future work we plan to account for large deformations of the organ and update the model accordingly. We plan to utilize computationally fast approaches to segment the dVRK tools from the images and avoid any obstructions to the overlaid stiffness map. Furthermore, as demonstrated by other researchers in this field, we believe a hybrid force-position controller can result in more accurate probing and hence better stiffness estimation. Finally, we plan to perform experiments on ex-vivo organs to asses the efficacy of the system in a realistic surgical setting.

\section*{Acknowledgment}
Special thanks to James Picard, Sasank Vishnubhatla, Peggy Martin and other colleagues from Biorobotics Lab, Carnegie Mellon University.

\bibliographystyle{IEEEtran}
\bibliography{arun}

\begin{thebibliography}{10}
\providecommand{\url}[1]{#1}
\csname url@samestyle\endcsname
\providecommand{\newblock}{\relax}
\providecommand{\bibinfo}[2]{#2}
\providecommand{\BIBentrySTDinterwordspacing}{\spaceskip=0pt\relax}
\providecommand{\BIBentryALTinterwordstretchfactor}{4}
\providecommand{\BIBentryALTinterwordspacing}{\spaceskip=\fontdimen2\font plus
\BIBentryALTinterwordstretchfactor\fontdimen3\font minus
  \fontdimen4\font\relax}
\providecommand{\BIBforeignlanguage}[2]{{%
\expandafter\ifx\csname l@#1\endcsname\relax
\typeout{** WARNING: IEEEtran.bst: No hyphenation pattern has been}%
\typeout{** loaded for the language `#1'. Using the pattern for}%
\typeout{** the default language instead.}%
\else
\language=\csname l@#1\endcsname
\fi
#2}}
\providecommand{\BIBdecl}{\relax}
\BIBdecl

\bibitem{jaydeeprmis09}
J.~Palep, ``Robotic assisted minimally invasive surgery,'' \emph{Journal of
  Minimal Access Surgery}, vol.~5, no.~1, pp. 1--7, Jan 2009.

\bibitem{garg2016tumor}
A.~Garg, S.~Sen, R.~Kapadia, Y.~Jen, S.~McKinley, L.~Miller, and K.~Goldberg,
  ``Tumor localization using automated palpation with gaussian process adaptive
  sampling,'' in \emph{CASE}.\hskip 1em plus 0.5em minus 0.4em\relax IEEE,
  2016, pp. 194--200.

\bibitem{li2017development}
L.~Li, B.~Yu, C.~Yang, P.~Vagdargi, R.~A. Srivatsan, and H.~Choset,
  ``Development of an inexpensive tri-axial force sensor for minimally invasive
  surgery,'' in \emph{In proceedings of the International Conference on
  Intelligent Robots and Systems}.\hskip 1em plus 0.5em minus 0.4em\relax IEEE,
  2017.

\bibitem{ayvali2016using}
E.~Ayvali, R.~A. Srivatsan, L.~Wang, R.~Roy, N.~Simaan, and H.~Choset, ``Using
  bayesian optimization to guide probing of a flexible environment for
  simultaneous registration and stiffness mapping,'' in \emph{ICRA}, no.
  10.1109/ICRA.2016.7487225, 2016, pp. 931--936.

\bibitem{ayvali2017utility}
E.~Ayvali, A.~Ansari, L.~Wang, N.~Simaan, and H.~Choset, ``Utility-guided
  palpation for locating tissue abnormalities,'' \emph{IEEE Robotics and
  Automation Letters}, vol.~2, no.~2, pp. 864--871, 2017.

\bibitem{hadi2018trajectory}
H.~Salman, E.~Ayvali, R.~A. Srivatsan, Y.~Ma, N.~Zevallos, R.~Yasin, L.~Wang,
  N.~Simaan, and H.~Choset, ``Trajectory-optimized sensing for active search of
  tissue abnormalities in robotic surgery,'' in \emph{Submitted to ICRA}.\hskip
  1em plus 0.5em minus 0.4em\relax IEEE, 2018.

\bibitem{srivatsan2016estimating}
R.~A. Srivatsan, G.~T. Rosen, F.~D. Naina, and H.~Choset, ``{Estimating SE(3)
  elements using a dual quaternion based linear Kalman filter},'' in
  \emph{Robotics : Science and Systems}, 2016.

\bibitem{srivatsanbingham2017}
R.~A. Srivatsan, M.~Xu, N.~Zevallos, and H.~Choset, ``{Bingham
  Distribution-Based Linear Filter for Online Pose Estimation},'' in
  \emph{Robotics : Science and Systems}, 2017.

\bibitem{kartikDynamic2017}
K.~Patath, R.~A. Srivatsan, N.~Zevallos, and H.~Choset, ``{Dynamic Texture
  Mapping of 3D models for Stiffness Map Visualization},'' in \emph{Workshop on
  Medical Imaging, IEEE/RSJ International Conference on Intelligent Robots and
  Systems}, 2017.

\bibitem{puangmali2008state}
P.~Puangmali, K.~Althoefer, L.~D. Seneviratne, D.~Murphy, and P.~Dasgupta,
  ``{State-of-the-art in force and tactile sensing for minimally invasive
  surgery},'' \emph{IEEE Sensors Journal}, vol.~8, no.~4, pp. 371--381, 2008.

\bibitem{tiwana2012review}
M.~I. Tiwana, S.~J. Redmond, and N.~H. Lovell, ``A review of tactile sensing
  technologies with applications in biomedical engineering,'' \emph{Sensors and
  Actuators A: physical}, vol. 179, pp. 17--31, 2012.

\bibitem{preetham2017online}
P.~Chalasani, L.~Wang, R.~Yasin, N.~Simaan, and H.~Taylor, Russel, ``Online
  estimation of organ geometry and tissue stiffness using continuous
  palpation,'' \emph{submitted to IEEE Robotics and Automation Letters}, 2017.

\bibitem{shuhaiber2004augmented}
J.~H. Shuhaiber, ``Augmented reality in surgery,'' \emph{Archives of surgery},
  vol. 139, no.~2, pp. 170--174, 2004.

\bibitem{teber2009augmented}
D.~Teber, S.~Guven, T.~Simpfend{\"o}rfer, M.~Baumhauer, E.~O. G{\"u}ven,
  F.~Yencilek, A.~S. G{\"o}zen, and J.~Rassweiler, ``Augmented reality: a new
  tool to improve surgical accuracy during laparoscopic partial nephrectomy?
  preliminary in vitro and in vivo results,'' \emph{European urology}, vol.~56,
  no.~2, pp. 332--338, 2009.

\bibitem{su2009augmented}
L.-M. Su, B.~P. Vagvolgyi, R.~Agarwal, C.~E. Reiley, R.~H. Taylor, and G.~D.
  Hager, ``{Augmented reality during robot-assisted laparoscopic partial
  nephrectomy: toward real-time 3D-CT to stereoscopic video registration},''
  \emph{Urology}, vol.~73, no.~4, pp. 896--900, 2009.

\bibitem{yamamoto2009tissue}
T.~Yamamoto, B.~Vagvolgyi, K.~Balaji, L.~L. Whitcomb, and A.~M. Okamura,
  ``Tissue property estimation and graphical display for teleoperated
  robot-assisted surgery,'' in \emph{ICRA}, 2009, pp. 4239--4245.

\bibitem{mckinley2015single}
S.~McKinley, A.~Garg, S.~Sen, R.~Kapadia, A.~Murali, K.~Nichols, S.~Lim,
  S.~Patil, P.~Abbeel, A.~M. Okamura \emph{et~al.}, ``A single-use haptic
  palpation probe for locating subcutaneous blood vessels in robot-assisted
  minimally invasive surgery,'' in \emph{CASE}.\hskip 1em plus 0.5em minus
  0.4em\relax IEEE, 2015, pp. 1151--1158.

\bibitem{naidu2017breakthrough}
A.~S. Naidu, M.~D. Naish, and R.~V. Patel, ``A breakthrough in tumor
  localization,'' \emph{IEEE Robotics \& Automation Magazine}, vol. 1070, no.
  9932/17, 2017.

\bibitem{trejos2009robot}
A.~L. Trejos, J.~Jayender, M.~Perri, M.~D. Naish, R.~V. Patel, and
  R.~Malthaner, ``Robot-assisted tactile sensing for minimally invasive tumor
  localization,'' \emph{The International Journal of Robotics Research},
  vol.~28, no.~9, pp. 1118--1133, 2009.

\bibitem{gotovos2013active}
A.~Gotovos, N.~Casati, G.~Hitz, and A.~Krause, ``Active learning for level set
  estimation,'' in \emph{IJCAI}, 2013, pp. 1344--1350.

\bibitem{nichols2015methods}
K.~A. Nichols and A.~M. Okamura, ``Methods to segment hard inclusions in soft
  tissue during autonomous robotic palpation,'' \emph{IEEE Transactions on
  Robotics}, vol.~31, no.~2, pp. 344--354, 2015.

\bibitem{liu2010rolling}
H.~Liu, D.~P. Noonan, B.~J. Challacombe, P.~Dasgupta, L.~D. Seneviratne, and
  K.~Althoefer, ``Rolling mechanical imaging for tissue abnormality
  localization during minimally invasive surgery,'' \emph{IEEE Transactions on
  Biomedical Engineering}, vol.~57, pp. 404--414, 2010.

\bibitem{goldman2013algorithms}
R.~E. Goldman, A.~Bajo, and N.~Simaan, ``Algorithms for autonomous exploration
  and estimation in compliant environments,'' \emph{Robotica}, vol.~31, no.~1,
  pp. 71--87, 2013.

\bibitem{howe1995remote}
R.~D. Howe, W.~J. Peine, D.~Kantarinis, and J.~S. Son, ``Remote palpation
  technology,'' \emph{IEEE Engineering in Medicine and Biology Magazine},
  vol.~14, no.~3, pp. 318--323, 1995.

\bibitem{srivatsan2016complementary}
R.~A. Srivatsan, E.~Ayvali, L.~Wang, R.~Roy, N.~Simaan, and H.~Choset,
  ``{Complementary Model Update: A Method for Simultaneous Registration and
  Stiffness Mapping in Flexible Environments},'' in \emph{ICRA}, 2016, pp.
  924--930.

\bibitem{marescaux2004augmented}
J.~Marescaux, F.~Rubino, M.~Arenas, D.~Mutter, and L.~Soler,
  ``Augmented-reality--assisted laparoscopic adrenalectomy,'' \emph{Jama}, vol.
  292, no.~18, pp. 2211--2215, 2004.

\bibitem{haouchine2014towards}
N.~Haouchine, J.~Dequidt, I.~Peterlik, E.~Kerrien, M.-O. Berger, and S.~Cotin,
  ``Towards an accurate tracking of liver tumors for augmented reality in
  robotic assisted surgery,'' in \emph{ICRA}, 2014, pp. 4121--4126.

\bibitem{besl1992method}
P.~J. Besl and N.~D. McKay, ``{Method for registration of 3-D shapes},'' in
  \emph{Robotics-DL tentative}.\hskip 1em plus 0.5em minus 0.4em\relax
  International Society for Optics and Photonics, 1992, pp. 586--606.

\bibitem{rusinkiewicz2001}
S.~Rusinkiewicz and M.~Levoy, ``{Efficient variants of the ICP algorithm},'' in
  \emph{Proceedings of 3rd International Conference on 3-D Digital Imaging and
  Modeling}.

\bibitem{billings2015iterative}
S.~D. Billings, E.~M. Boctor, and R.~H. Taylor, ``{Iterative most-likely point
  registration (IMLP): A robust algorithm for computing optimal shape
  alignment},'' \emph{PloS one}, vol.~10, no.~3, p. e0117688, 2015.

\bibitem{moghari2007point}
M.~H. Moghari and P.~Abolmaesumi, ``{Point-based rigid-body registration using
  an unscented Kalman filter},'' \emph{IEEE Transactions on Medical Imaging},
  vol.~26, no.~12, pp. 1708--1728, 2007.

\bibitem{kazanzides2014open}
P.~Kazanzides, Z.~Chen, A.~Deguet, G.~S. Fischer, R.~H. Taylor, and S.~P.
  DiMaio, ``An open-source research kit for the da vinci{\textregistered}
  surgical system,'' in \emph{ICRA}.\hskip 1em plus 0.5em minus 0.4em\relax
  IEEE, 2014, pp. 6434--6439.

\bibitem{horn1987}
B.~K. Horn, ``Closed-form solution of absolute orientation using unit
  quaternions,'' \emph{JOSA A}, vol.~4, no.~4, pp. 629--642, 1987.

\end{thebibliography}

\end{document}